\documentclass[conference]{IEEEtran}
\IEEEoverridecommandlockouts
\usepackage{cite}
\usepackage{amsmath,amssymb,amsfonts}
\usepackage{algorithmic}
\usepackage{graphicx}
\usepackage{textcomp}
\usepackage{xcolor}
\usepackage{multirow}
\usepackage{flushend}
\usepackage{hyperref}

\def\BibTeX{{\rm B\kern-.05em{\sc i\kern-.025em b}\kern-.08em
    T\kern-.1667em\lower.7ex\hbox{E}\kern-.125emX}}
\begin{document}

\title{Evaluation Framework for AI-driven Molecular Design of Multi-target Drugs: Brain Diseases as a Case Study
\thanks{© 2024 IEEE. Personal use of this material is permitted. Permission from IEEE must be obtained for all other uses, in any current or future media, including reprinting/republishing this material for advertising or promotional purposes, creating new collective works, for resale or redistribution to servers or lists, or reuse of any copyrighted component of this work in other works.
DOI: \href{https://doi.org/10.1109/CEC60901.2024.10611839}{10.1109/CEC60901.2024.10611839}}}

\author{
\IEEEauthorblockN{Arthur Cerveira\textsuperscript{1,2}, Frederico Kremer\textsuperscript{2,3}, Darling Lourenço\textsuperscript{3}, Ulisses B. Corrêa\textsuperscript{1,2}}
\IEEEauthorblockA{
\textsuperscript{1}Programa de Pós-graduação em Computação (PPGC)\\
\textsuperscript{2}Hub de Inovação em Inteligência Artificial (H2IA)\\
\textsuperscript{3}Omixlab, \\
Centro de Desenvolvimento Tecnológico (CDTec),\\ Universidade Federal de Pelotas (UFPel),\\ Brazil\\
\{aacerveira@inf.ufpel.edu.br, fred.s.kremer@gmail.com, darlinglourenco@gmail.com, ulisses@inf.ufpel.edu.br\}
}
}
\maketitle

\begin{abstract}
The widespread application of Artificial Intelligence (AI) techniques has significantly influenced the development of new therapeutic agents. These computational methods can be used to design and predict the properties of generated molecules. Multi-target Drug Discovery (MTDD) is an emerging paradigm for discovering drugs against complex disorders that do not respond well to more traditional target-specific treatments, such as central nervous system, immune system, and cardiovascular diseases. Still, there is yet to be an established benchmark suite for assessing the effectiveness of AI tools for designing multi-target compounds. Standardized benchmarks allow for comparing existing techniques and promote rapid research progress. Hence, this work proposes an evaluation framework for molecule generation techniques in MTDD scenarios, considering brain diseases as a case study. Our methodology involves using large language models to select the appropriate molecular targets, gathering and preprocessing the bioassay datasets, training quantitative structure-activity relationship models to predict target modulation, and assessing other essential drug-likeness properties for implementing the benchmarks. Additionally, this work will assess the performance of four deep generative models and evolutionary algorithms over our benchmark suite. In our findings, both evolutionary algorithms and generative models can achieve competitive results across the proposed benchmarks.
\end{abstract}

\begin{IEEEkeywords}
molecular design, multi-target drug discovery, evolutionary algorithms, deep generative models, de novo design
\end{IEEEkeywords}

\section{Introduction}




Drug discovery research is a multidisciplinary field that aims to identify and develop novel therapeutic compounds for the treatment of diseases~\cite{blass2021drugdiscovery}. The Drug Discovery Pipeline (DDP) is a costly and time-consuming process  
to advance from hit molecule candidates to clinical-trial-ready compounds. Computational and Artificial Intelligence (AI) approaches are frequently employed in the early stages of the DDP to accelerate the discovery of candidate compounds. These computational methods are also called \textit{in-silico} drug discovery techniques~\cite{chang2022insilico}.




Traditional \textit{in-silico} strategies rely on High-Throughput Screening (HTS) techniques to identify molecules that optimally satisfy a desired molecular profile in a pre-generated compound library~\cite{chang2022insilico}. This approach is limited to identifying drug candidates from known compounds, representing a tiny fraction of the drug-like chemical space~\cite{polishchuk2013chemspace}. The \textit{de Novo} Molecular Design (dNMD) approach utilizes AI models and algorithms to generate novel compounds from scratch, optimizing them to meet a desired molecular profile~\cite{meyers2021denovo}.




The one-target-one-disease paradigm has been the central concept behind discovering new drugs. This approach involves designing ligands that act on a single target associated with each disease~\cite{franco2013tsdtomtd}. Although many successful drugs have arisen from this principle, there is a growing conviction that the modulation of multiple targets can be advantageous when treating multifactorial diseases~\cite{morphy2005dml}. For example, clinicians already employ multi-target molecular entities to manage various hematologic malignancies and solid tumors~\cite{gentile2017kinase}. 

Multi-target Drug Discovery (MTDD) is the field responsible for developing new molecules with multiple targets. Multi-Target Drugs (MTDs) can present some advantages over Target-Specific Drugs (TSDs), such as improved efficacy due to synergistic effects, reduced adverse reactions, fewer drug-drug interactions, and lower risk of toxicity~\cite{wang2019mtddalzheimer}. However, the design of novel MTDs also comprises many challenges, as these drugs may interact with unwanted targets, which can cause undesired side effects~\cite{morphy2005dml}.









The rise of AI-based dNMD techniques for molecule generation motivated the development of Guacamol, a benchmark suite for assessing and comparing dNMD models and algorithms~\cite{brown2019guacamol}. The Guacamol benchmarks can be grouped into two main use cases: (1) distribution-learning benchmarks, where the model must generate new molecules following the same chemical distribution of its training set, and (2) goal-directed benchmarks, where the model generates the best molecules optimized to satisfy a predefined goal.

While distribution-learning benchmarks do not distinguish single- and multi-target scenarios, Guacamol goal-directed benchmarks mostly lean towards TSD-based goals. Currently, there is no established goal-directed benchmark for assessing the effectiveness of dNMD methods for MTDD tasks. To address this gap, we introduce a new disease-guided evaluation framework for MTDD use cases, enabling comparative analysis of 
molecule generation approaches and techniques.



As the primary contributions of this work, we propose:

\begin{enumerate}
    \item a new evaluation methodology to assess AI-driven molecular design strategies in MTDD scenarios;
    \item a comparison between molecular generation techniques in our evaluation framework, considering brain diseases as a case study.
\end{enumerate}
As the specific objectives, we design disease-guided scoring functions to assess the response against the specified targets and other relevant drug-likeness properties. We also introduce a target selection algorithm that leverages the application of Large Language Models (LLMs) to identify an appropriate combination of protein targets for each disease. To design the benchmark scoring functions, we aim to extend the novel BioAssays Model Builder (Bambu) tool~\cite{guidotti2023bambu} by introducing new practical benchmarks for assessing Quantitative Structure-Activity Relationship (QSAR) models in lead optimization tasks.

\section{Background and Related Work}

\bigskip


\subsection{Multi-target Drugs for Complex Diseases}





MTDs, also known as designed multiple ligands, have arisen as a promising alternative in drug discovery for developing treatments for multi-factor disorders, such as central nervous system diseases, cancer, immune diseases, and cardiovascular diseases~\cite{viana2018mtdd}. Different from TSDs, these therapeutic agents aim to modulate multiple targets simultaneously, addressing the complex nature of many physiological processes. The polypharmacology concept of modulating multiple targets comes from the understanding that diseases often result from a network of interconnected pathways rather than a single molecular target~\cite{franco2013tsdtomtd}. 






Still, designing effective MTDs remains a very challenging task. The complexity of biological systems and the potential for off-target effects require a deep understanding of the interactions between different targets. Also, identifying compounds capable of modulating multiple targets while having other essential drug-likeness features is a complex multi-objective optimization problem~\cite{nicolaou2013moo}. Despite these challenges, the advantages of well-designed MTDs over TSDs are significant, including achieving synergistic effects by targeting multiple disease-related pathways simultaneously, enhancing therapeutic efficacy, and minimizing the likelihood of drug resistance~\cite{viana2018mtdd}.


Target selection is decisive when developing new multi-target therapeutic compounds and requires understanding the determined disease's physiopathology~\cite{viana2018mtdd}. When developing MTDs, the simultaneous modulation of multiple targets should result in synergistic therapeutic effects for treating the disease. However, not all target combinations are suitable for being tackled simultaneously, as some sub-pathologies might occur in different stages of a disease~\cite{benek2020alzheimers}.

\subsection{Assessment of de Novo Molecular Design Strategies}





The two main evaluation approaches for dNMD techniques are distribution-learning and goal-directed generation. Distribution-learning tasks assess the quality and diversity of the generated molecules compared with the model's training set. In goal-directed generation, the model must generate the most optimized molecules according to a desired molecular profile. Distribution-learning benchmarks assess the entire generated molecule set, while goal-directed benchmarks score each molecule individually.

Guacamol and  Molecular Sets (MOSES) are the most popular benchmark suites for assessing the capability of dNMD in reproducing the distribution of their training set. Both benchmark suites include the validity, uniqueness, and novelty metrics to evaluate the generated molecules and the Fréchet ChemNet Distance metric to compare them with the training data distribution. MOSES also includes other benchmarks based on molecule filtering, various similarity measures, internal diversity, and property distribution. Guacamol includes the Kullback–Leibler (KL) divergence as a distribution-learning benchmark as well.

Guacamol implements diverse benchmarks for goal-directed molecule generation tasks, evaluating their conformity to a specified property profile or Multi-Property Objective (MPO). dNMD models must be capable of iteratively optimizing the generated molecules without explicit knowledge of the scoring function. The benchmarks assess molecular features such as structural characteristics, physicochemical properties, and similarity or dissimilarity to other molecules. The MPO benchmarks often aggregate the individual scoring functions through their geometric mean.


Recent research discusses the feasibility of realistically validating molecular generative models.~\cite{handa2023dnmdchallenges} highlight the difficulty of dNMD models in recovering middle/late-stage project compounds using information from early-stage compounds. This view is a reasonable argument for a lead optimization task, where we want to optimize a hit molecule towards a specific goal. Still, researchers demonstrated how generative models can achieve early-stage hit rate on par with traditional molecule selection approaches guided by an expert medicinal chemist~\cite{korshunova2022nature}.

\bigskip

\subsection{Artificial Intelligence Techniques for Molecular Design}

Using AI algorithms and models to design novel molecules is a growing paradigm in the drug discovery field. Although HTS approaches are still widely used for identifying candidate drugs, AI-driven dNMD approaches have shown promising results in designing reasonable early-stage compounds with therapeutical properties~\cite{meyers2021denovo}. Evolutionary Algorithms (EA) and Deep Generative Models (DGM) are two of the most common strategies for molecule generation.




AI-driven dNMD strategies must be capable of generating and scoring new molecules, optimizing their exploration based on the corresponding scoring outcomes. This way, every iteration of the molecule generation process will achieve increasingly tailored compounds to meet a desired molecular profile. The scoring function may comprise multiple objectives and criteria that should be optimized simultaneously.




\subsection{QSAR Models for Drug Discovery}


Quantitative Structure-Activity Relationship (QSAR) modeling is an \textit{in silico} technique used to estimate a molecule's properties based on its structure~\cite{chang2022insilico}. HTS and dNMD approaches can use QSAR models to score candidate molecules in drug discovery. Building and training QSAR models typically involve using statistical or Machine Learning (ML) algorithms and training data from \textit{in vitro} tested molecules. These models can be used to predict the biological activity of a molecule against a disease protein target, for example~\cite{guidotti2023bambu}. 

\section{Evaluation Framework for Multi-target Drug Discovery}



Our evaluation framework follows a goal-directed generation approach, where the dNMD models and algorithms must optimize the generated molecules towards a predefined goal. Each benchmark implementation is guided by the pharmacological profile of the determined disease, considering its known targets and other specific drug-likeness features required for functional active compounds. Our methodology uses available FDA-approved drugs and findings from scientific research as the primary sources of this information.




It is crucial to highlight that drug discovery is not ligand discovery, and the design of new therapeutic compounds must consider more than their effectiveness against the defined targets~\cite{handa2023dnmdchallenges}. The molecular design stage of the DDP consists of a Multi-Property Optimization (MPO) problem, optimizing simultaneously for target activity, physicochemical and pharmacokinetics properties, synthetic accessibility, and so forth~\cite{nicolaou2013moo}. Accordingly, we must incorporate these aspects into the benchmark scoring functions to design a robust evaluation framework.



As a case study, our work introduces three disease-guided benchmarks to evaluate AI-driven dNMD strategies. The conditions defined for this evaluation are Alzheimer's Disease (AD), Schizophrenia, and Parkinson's Disease (PD). We identified the primary targets for each disease based on available and potential drug candidates and information derived from the literature. We also identified suitable physicochemical and pharmacokinetics features for the evaluation. Likewise, the synthetic accessibility of the generated molecule will also impact the benchmark scoring functions. Still, the aspects considered for this evaluation do not represent an exhaustive list of required characteristics for therapeutic agents. This section will explain the scoring functions defined for each considered condition.








\subsection{Alzheimer's Disease MPO Benchmark}


AD is a multifactorial neurodegenerative disease characterized by pathological proteins, impaired neurotransmission, increased oxidative stress, and microglia-mediated neuroinflammation~\cite{knopman2021alzheimer}. Recent research indicates MTDD as a promising approach for treating AD by simultaneously acting on multiple sub-pathologies. Currently, acetylcholinesterase (AChE) inhibitors and N-methyl-D-aspartate receptor (NMDAR) antagonists are the only FDA-approved drugs in clinical use for AD. MTDD strategies resulted in the development of two clinical candidates for AD; however, neither completed the validation stage in clinical trials~\cite{benek2020alzheimers}.


The properties considered for the AD MPO benchmark are the activity against the associated pathological proteins, the ability to cross the Blood-Brain Barrier (BBB)~\cite{meng2021bbb}, the physicochemical properties of the Central Nervous System Multiparameter Optimization (CNS MPO) from Pfizer~\cite{zoran2015cnsmpo}, and the synthetic accessibility score to discriminate feasible molecules from infeasible ones. The molecular targets considered are AChE and monoamine oxidase B (MAO-B), both targeted by MTDs in recent clinical trials~\cite{weinreb2012ladostigil}. Although the amyloid-$\beta$ peptide (A$\beta$) is also considered a crucial player in AD progression, the simultaneous targeting of neurotransmission and A$\beta$ represents an inappropriate drug design as those two events take place at different stages of the disease~\cite{benek2020alzheimers}.




\subsection{Schizophrenia MPO Benchmark}

Schizophrenia is a complex neuropsychiatric disorder resulting from disturbances in neurotransmission involving a significant number of receptors and enzymes, primarily within the dopaminergic, glutamatergic, serotoninergic, and adrenergic systems~\cite{maletic2017schizophrenia}. TSDs for schizophrenia have been performing below expectations, and MTDD strategies have been emerging as an alternative for treating this disease~\cite{kondej2018schizophrenia}. While the dopaminergic hypothesis remains the most widely accepted explanation of the disease, novel strategies in drug discovery against schizophrenia focus on targets beyond this concept~\cite{lau2013dopamine}.

For the schizophrenia MPO benchmark, we also considered the molecule response against the associated receptors, the ability to cross the BBB, the CNS MPO for its physicochemical properties, and the synthetic accessibility score. The determined targets are the dopamine D2 receptor (D2R) and the serotonin 5-hydroxytryptamine 2A receptor (5-HT2AR) — known pharmacologically active targets of the available FDA-approved drugs for schizophrenia~\cite{kondej2018schizophrenia}.


\subsection{Parkinson's Disease MPO Benchmark}


PD is a neurodegenerative disease characterized by the progressive degeneration of dopaminergic neurons, leading to motor and non-motor complications. 
While the molecular mechanisms of PD are not fully understood, studies suggest that factors such as mitochondrial dysfunction, oxidative stress, neuroinflammation, and the accumulation of alpha-synuclein contribute to the pathogenesis of Parkinson's~\cite{dongchen2023parkinson}. Current therapeutic strategies mainly focus on alleviating symptoms through dopaminergic medications; however, disease-modifying treatments are still in the experimental stages~\cite{antonelli2014parkinson}. 

The MPO benchmark for PD includes evaluating a compound's ability to modulate the defined targets, penetrate the BBB, adhere to the CNS MPO guidelines, and achieve a synthetic accessibility score indicative of feasible drug development. The benchmark targets two dopamine receptors, the dopamine D2 receptors (D2R) and dopamine D3 receptors (D3R), essential proteins in the dopaminergic system, to modulate neurotransmission and address motor symptoms of the disease~\cite{antonelli2014parkinson}.

\begin{figure*}[t]
    \centering
\includegraphics[width=500px]{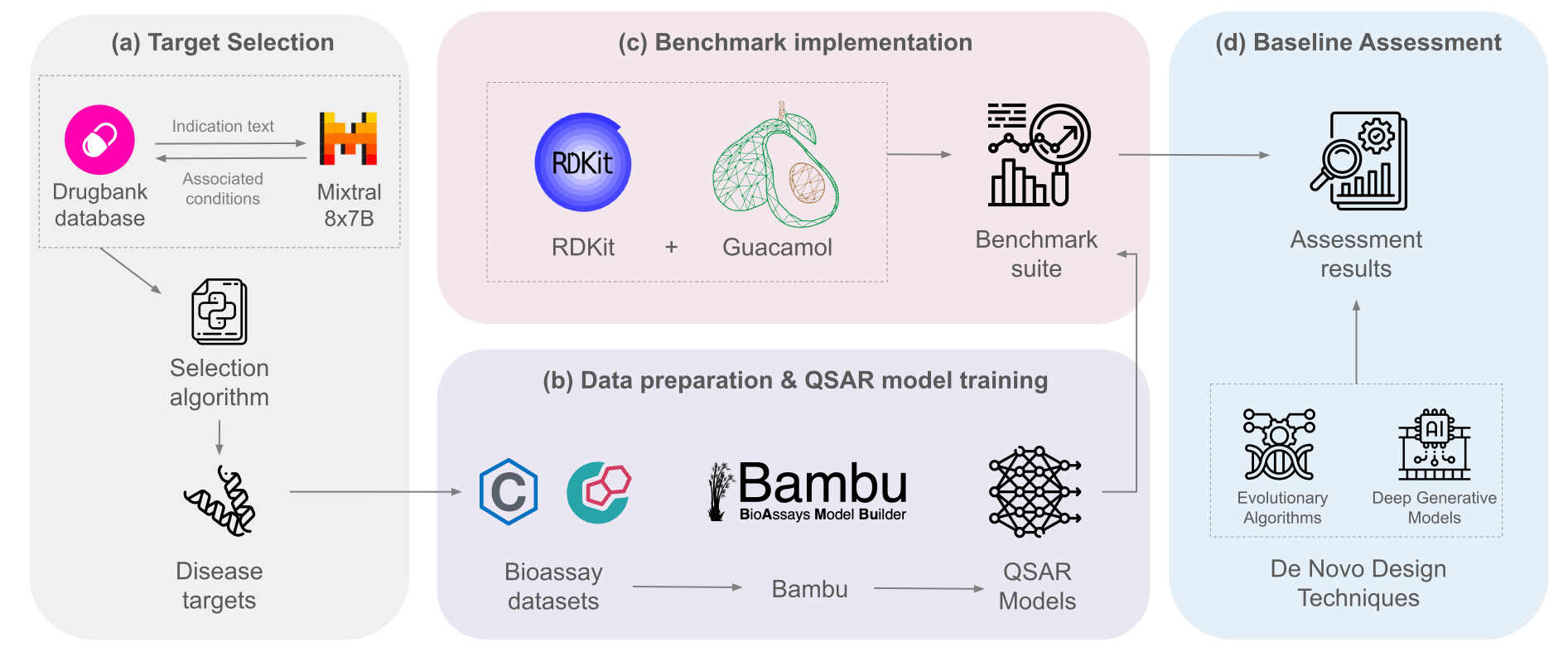}
    \caption{Proposed methodology illustrating (a) target selection, (b) data preparation and QSAR model training, (c) benchmark implementation, and (d) baseline assessment stages.}
    \label{fig:methodology}
\end{figure*}

\section{Methodology}

The proposed methodology consists of (1) selecting the appropriate disease targets, (2) preparing the datasets, (3) training the QSAR models, (4) implementing the disease-guided benchmarks, and (5) assessing the dNMD techniques over our benchmark suite. Figure \ref{fig:methodology} depicts the methodology diagram. The code and data for reproducing the methodology is available at GitHub\footnote{https://github.com/arthurcerveira/MTDD-Evaluation-Framework}.





\subsection{Target Selection}










Our target selection methodology employs the DrugBank database version 5.1.10, containing data on 15,235 therapeutic compounds, including their indication text and protein targets \cite{knox2024drugbank}. To determine the targets for each benchmark, we used an LLM to perform a structured extraction of the associated disease for each small molecule drug in the database. 
Then, we computed the co-occurrence matrix of the protein targets for each disease to identify the most common target combinations. This co-occurrence matrix is normalized according to the highest co-occurrence value, and a greedy algorithm iteratively selects the best combination of targets until the maximum co-occurrence value reaches a predefined threshold.

We employed the Mixtral 8x7B LLM from Mistral AI to perform the information extraction step, as it achieves state-of-the-art results among the available open-source models on human evaluation benchmarks~\cite{jiang2024mixtral}. The associated conditions were classified into Alzheimer's, Schizophrenia, Parkinson's, and Others, based on their indication text. We chose 0.7 as the greedy algorithm threshold to select the combination of targets, based on empirical analysis indicating its reliable performance. Table \ref{tab:target-selection} presents the targets selected for each condition. The AChE protein selected for AD is a type of Cholinesterase, so both proteins represent the same target. For this reason, the MAO-B protein was included as a second target for our Alzheimer's MPO benchmark.

\begin{table}[ht]
\caption{Target selection results for each associated condition.}
\label{tab:target-selection}
{
\centering
\begin{tabular}{l|l}
\textbf{Condition} & \textbf{Selected Targets} \\ \hline
\multirow{2}{*}{Alzheimer's Disease} & Cholinesterase \\ 
 & Acetylcholinesterase \\ \hline
\multirow{2}{*}{Schizophrenia} & Dopamine D2 receptor \\ 
 & 5-hydroxytryptamine 2A receptor \\ \hline
\multirow{2}{*}{Parkinson's Disease} & Dopamine D2 receptor \\ 
 & Dopamine D3 receptors
\end{tabular}\par
}
\end{table}

\subsection{Data Preparation}

Training QSAR models for predicting activity against a target requires quality datasets containing a set of molecules and their measured \textit{in vitro} activity. After identifying the targets determined for each disease MPO, we used the download function from the Bambu API to collect and aggregate most of the bioassay datasets from the PubChem platform\footnote{https://pubchem.ncbi.nlm.nih.gov/}. We enriched the datasets with information from ChEMBL\footnote{https://www.ebi.ac.uk/chembl/} when the required data was unavailable at PubChem.

Some AI-driven dNMD models and algorithms also require a drug-like molecule database for generating novel therapeutic compounds. We derived this database from the ChEMBL 24 compound library, which only contains synthesizable molecules. Following the Guacamol dataset generation methodology, we processed this database by (1) removing salts, (2) neutralizing their charges, (3) removing molecules with $>100$ SMILES (Simplified Molecular-Input Line-Entry System) representation string length, and (4) filtering only the molecules that exclusively contain the elements H, B, C, N, O, F, Si, P, S, Cl, Se, Br, and I. We also removed all molecules with a Tanimoto ECFP4 similarity higher than 0.323 compared to any of the molecules used as the basis for our scoring functions (from the drugs identified for each disease in the target selection phase), guaranteeing a fair comparison between the dNMD strategies.

\subsection{QSAR Model Training}

ML and statistical algorithms power the QSAR models employed in the proposed scoring functions. Our model training methodology uses Bambu  — an AutoML package for QSAR models developed on top of the RDKit\footnote{https://www.rdkit.org} and FLAML\footnote{https://microsoft.github.io/FLAML} libraries. The Bambu API offers an interface to download and preprocess bioassay datasets from PubChem, and train the QSAR model by evaluating various ML algorithms and hyperparameter setups. 


The model evaluation and data split employ the Lo-Hi Splitter, a new practical benchmark method for drug discovery~\cite{steshin2023lohi}. More specifically, we evaluated and selected the QSAR models on the Lo task, as it simulates a real-world lead optimization scenario associated with goal-directed molecular design tasks. On this data split, the models should predict properties of minor modifications of molecules with known activity. We also extended the Bambu model inference API with a faster prediction implementation through vectorization techniques.




To preprocess the datasets, we computed the molecule's circular fingerprints using the Morgan algorithm with a radius of 2 and a fingerprint size of 1024 bits. The ML classification algorithms assessed in the AutoML pipeline are the random forest, extra trees, decision tree, support vector machine, gradient boosting, and multilayer perceptron, optimizing their hyper-parameter configurations through random search for a time budget of 3600 seconds. For the Lo dataset split, we considered the parameters 0.323 as similarity threshold (same value defined for filtering out molecules in the data preparation step), 5 for minimum cluster size, and 50 for the maximum cluster size.

\subsection{Benchmark Implementation}


QSAR models were employed to predict the molecule activity against the defined targets, where the activity can describe inhibition, antagonism, or agonism activity according to the training dataset. In this context, the model returns a score between 0 and 1, with higher scores indicating increased molecular activity. Those scores are aggregated through a geometric mean operation to penalize molecules with zero predicted activity. 
Previous works have also considered mean binding affinity towards multiple proteins as a scoring function for multi-target molecular design~\cite{chebrolu2023mtdd}.



The two metrics determined to evaluate the physicochemical properties are the CNS MPO and the ability to cross the BBB. The RDKit package provides the operations necessary to implement the CNS MPO scoring functions. We designed the BBB scorer as a QSAR model trained on the B3DB dataset~\cite{meng2021b3bd}, predicting the molecule's ability to cross the BBB based on its features. Each physicochemical property is treated as an individual scoring function for the MPO benchmark.


We use the SAScore as our synthetic accessibility scoring function, calculated as a sum of fragment scores and a complexity penalty~\cite{ertl2009sascore}. A recent study demonstrated this method as the most reliable synthetical accessibility scorer, with an AUC and accuracy over 0.81 in retrosynthetic planning tasks~\cite{skoraczynski2023saassessment}. SAScore can discriminate feasible molecules from infeasible ones in the majority of cases. Besides the improvements in the accuracy of scores, current ML techniques do not entirely replace a human mind in the retrosynthesis planning process~\cite{skoraczynski2023saassessment}. We designed the scoring function to output values between 0 and 1, where higher values indicate increased synthetic accessibility scores.

The disease MPO scoring function combines individual scorer outputs using a geometric mean operation, returning a final score ranging from 0 to 1 for each molecule. This final score estimates the molecule's potential to advance as an early-stage candidate for MTD against the specified disease. Assessing the scores obtained by the most optimized molecules generated by the dNMD strategy provides a meaningful metric for quantifying and comparing its performance in MTDD scenarios.

\subsection{Assessing the Baseline Techniques}

This work assesses four dNMD models and algorithms in our proposed evaluation framework to compare and evaluate these strategies' capability to generate novel multi-target therapeutic compounds. We can classify the considered approaches into two groups: DGM-based and EA-based methods. For DGM-based strategies, we evaluate two LSTM models: one based on the hill-climb algorithm for optimization (LSTM-HC) \cite{segler2018lstmhc} and another based on the reinforcement learning concept of proximal policy optimization (LSTM-PPO) \cite{neil2018lstmppo}. We considered two other EA-based strategies for our analysis: a SMILES genetic algorithm (SMILES GA) \cite{naruki2018smilesga} and a graph-based genetic algorithm (Graph GA) \cite{jensen2019graphga}. As baselines, we also selected the highest-scoring molecule from our processed dNMD dataset for each benchmark. We considered the hyperparameter setup described in~\cite{brown2019guacamol} for a fair comparison between the dNMD techniques.

Both LSTM models were optimized for 20 epochs in each benchmark assessment. The LSTM-HC was fine-tuned over the top 512 molecules from a pool of 1024 sampled molecules at each training iteration. The LSTM-PPO reward function also incorporates an entropy term to promote diversification and a KL divergence term to mitigate deviation from the ChEMBL chemical space. For the EA-based algorithms, we considered 1000 generations, with a population size of 100 and 200 offsprings in each generation. The Graph GA was configured with a mutation rate of 0.01 and a patience level of 5 generations. Due to a slower convergence, we set a higher patience level of 50 generations for the SMILES GA to ensure the generation of optimized molecules.


\section{Results and Discussion}


\subsection{QSAR Models Evaluation}

The QSAR models were trained through an AutoML workflow that evaluated the performance of various ML algorithms and hyperparameter configurations. We compared the various training iterations on the Lo benchmark task to evaluate and select the best model, which simulates a practical lead optimization scenario. Table \ref{tab:qsar-evaluation} presents the chosen algorithm and performance metrics for the BBB and each target bioassay datasets.

\begin{table}[!ht]
\caption{Results for the QSAR models on the Lo benchmark.}
\label{tab:qsar-evaluation}
{
\centering
\begin{tabular}{l|ccc}
\textbf{Target} & \textbf{Accuracy} & \textbf{ROC-AUC} & \textbf{Model} \\ \hline
AChE     & 0.575             & 0.567            & Gradient Boosting  \\
MAO-B    & 0.625             & 0.614            & Gradient Boosting \\
D2R      & 0.616             & 0.587            & Decision Tree  \\
5-HT2AR  & 0.556             & 0.541            & Decision Tree  \\
D3R      & 0.616             & 0.615            & Random Forest  \\
BBB      & 0.640             & 0.626            & Random Forest 
\end{tabular}\par
}
\end{table}

The metrics considered for evaluating the QSAR models are accuracy and ROC-AUC. Overall, there is a low variation in the performance of models for each target, achieving similar outcomes for both metrics. All the selected models are tree-based ML algorithms, which match results achieved in the other works training QSAR models for predicting target modulation with Bambu~\cite{guidotti2023bambu}. Tree-based algorithms can capture nonlinear relationships between the features and the target variable, making them suitable for representing complex decision boundaries. Still, for the D2R and 5-HT2AR datasets, classic decision tree models outperformed more complex ML algorithms for our evaluation.

\subsection{Disease-guided Benchmarks}

For our brain disease case study, we evaluated each dNMD method on all proposed benchmarks. We displayed the results in Table \ref{tab:mpo-benchmarks} with the scores achieved for each scoring function and the final result. The final score represents the benchmark result aggregating the individual properties scores for the MPO problem. The Best of Dataset represents an appropriate baseline for comparing the dNMD strategies with the more traditional HTS approach, as it identifies the highest-scoring molecules in the filtered SMILES dataset.

\begin{table*}[!t]
\caption{Results for the MPO benchmarks in the brain diseases case study. Average score and standard deviation across the top 100 molecules for individual scoring functions.}
\label{tab:mpo-benchmarks}
{
\centering
\begin{tabular}{ll|ccccc}
\textbf{Benchmark} & \textbf{Metric} & \textbf{LSTM-PPO} & \textbf{LSTM-HC} & \textbf{SMILES GA} & \textbf{Graph GA} & \textbf{Best of dataset} \\ \hline
\multirow{5}{*}{\begin{tabular}[c]{@{}l@{}}Alzheimer's\\ Disease\\ MPO\end{tabular}} & Score & 0.8527 & 0.8820 & 0.8766 & \textbf{0.8821} & 0.8743 \\
 & Target Response & 0.692 ± 0.02 & 0.709 ± 0.01 & 0.710 ± 0.01 & 0.712 ± 0.005 & 0.698 ± 0.02 \\
 & Blood-Brain Barrier & 0.857 ± 0.05 & 0.955 ± 0.02 & 0.934 ± 0.02 & 0.957 ± 0.015 & 0.924 ± 0.03 \\
 & CNS MPO & 0.971 ± 0.04 & 0.990 ± 0.01 & 0.983 ± 0.02 & 0.992 ± 0.01 & 0.977 ± 0.03 \\
 & Synthetic Accessibility & 0.851 ± 0.03 & 0.886 ± 0.02 & 0.877 ± 0.02 & 0.881 ± 0.015 & 0.886 ± 0.02 \\ \hline
\multirow{5}{*}{\begin{tabular}[c]{@{}l@{}}Schizophrenia\\ MPO\end{tabular}} & Score & 0.8486 & \textbf{0.8791} & 0.8663 & 0.8741 & 0.8639 \\
 & Target Response & 0.670 ± 0.02 & 0.697 ± 0.01 & 0.672 ± 0.02 & 0.693 ± 0.01 & 0.669 ± 0.02 \\
 & Blood-Brain Barrier & 0.851 ± 0.04 & 0.960 ± 0.01 & 0.931 ± 0.02 & 0.943 ± 0.02 & 0.919 ± 0.03 \\
 & CNS MPO & 0.967 ± 0.04 & 0.987 ± 0.02 & 0.979 ± 0.03 & 0.988 ± 0.02 & 0.977 ± 0.03 \\
 & Synthetic Accessibility & 0.863 ± 0.03 & 0.886 ± 0.015 & 0.886 ± 0.025 & 0.882 ± 0.01 & 0.892 ± 0.02 \\ \hline
\multirow{5}{*}{\begin{tabular}[c]{@{}l@{}}Parkinson's\\ Disease\\ MPO\end{tabular}} & Score & 0.8371 & 0.8731 & 0.8598 & \textbf{0.8738} & 0.8588 \\
 & Target Response & 0.643 ± 0.03 & 0.676 ± 0.01 & 0.658 ± 0.01 & 0.682 ± 0.01 & 0.655 ± 0.02 \\
 & Blood-Brain Barrier & 0.829 ± 0.05 & 0.950 ± 0.02 & 0.923 ± 0.03 & 0.959 ± 0.02 & 0.918 ± 0.03 \\
 & CNS MPO & 0.965 ± 0.03 & 0.986 ± 0.02 & 0.980 ± 0.03 & 0.990 ± 0.01 & 0.974 ± 0.03 \\
 & Synthetic Accessibility & 0.857 ± 0.03 & 0.894 ± 0.01 & 0.887 ± 0.02 & 0.885 ± 0.01 & 0.891 ± 0.02
\end{tabular}\par
}
\end{table*}


The Graph GA technique narrowly outperformed other methods for AD and PD MPO benchmarks, while LSTM-HC achieved the best Schizophrenia MPO results. Both EAs and DGMs achieved competitive results on the individual scoring functions. Most dNMD strategies outperformed the Best of Dataset in the final score as well, highlighting their efficacy in generating molecules optimized for specific profiles compared to the HTS approach. One potential explanation for HTS outperforming LSTM-PPO is that the latter does not start from a population of optimized molecules, so it might require further optimization for improved results.

Still, on this model-based assessment, our evaluation framework will only be as reliable as the models used on the scoring functions, and this prediction noise might be propagated when aggregating the individual metrics into a final score. The assessed models and algorithms  consider only the final aggregated score to optimize the molecule generation process. Thus, developing molecular design strategies that leverage the impact of individual scoring functions represents a promising approach to improving performance over this benchmark suite and advancing the MTDD field.

\section{Conclusions}


This work proposed a novel evaluation framework for AI-driven molecular design methods based on MTDD scenarios. Our benchmarking methodology assesses crucial aspects of the DDP, such as target modulation, physicochemical properties, and synthetic accessibility. As a case study, we implemented a set of benchmarks guided by the pharmacological profile of the brain diseases AD, Schizophrenia, and PD. Among the assessed dNMD strategies, the Graph GA and LSTM-HC techniques achieved the best scores across all three benchmarks, with both EA- and DGM-based approaches achieving competitive results in individual scoring functions. The presented framework has the potential for extension to various diseases and relevant properties, including aspects related to compound safety, toxicity prediction, and molecule promiscuity. Our methodology proves practical for assessing dNMD strategies and developing potential early-stage therapeutic compounds. Additionally, the insights present in this work can guide future strategies for designing dNMD techniques directed to MTDD use cases.

\section*{Acknowledgment}

We gratefully acknowledge the support of NVIDIA for donating the Titan X Pascal GPU used for this research. This study was financed in part by the Coordenação de Aperfeiçoamento de Pessoal de Nível Superior – Brazil (CAPES) – Finance Code 001, and FAPERGS - Brazil, Award Agreement 22/2551-0000598-5.


\bibliographystyle{IEEEtran}
\bibliography{custom}


\end{document}